\documentclass[11pt]{article}

\usepackage[preprint]{acl}

\usepackage{times}
\usepackage{pgfplots}
\usepackage{latexsym}
\usepackage{multirow}
\usepackage{amsmath}
\usepackage{linguex}
\usepackage{cgloss4e}

\usepackage[T1]{fontenc}

\usepackage[utf8]{inputenc}

\usepackage{microtype}

\usepackage{inconsolata}

\usepackage{graphicx}
\usepackage{tikz, tikz-qtree}

\usepackage{linguex}

\title{A graph-based analysis of semantic types and coercion in contextualized word embeddings}

\author{Long Chen ~~~~ Deniz Ekin Yava\c{s} \\
  Heinrich Heine University Düsseldorf \\
  \texttt{\{chen.long,deniz.yavas\}@hhu.de} 
  }

\newcommand{\hide}[1]{}
\newcommand{\type}[1]{\textit{#1}}

\begin{document}
\maketitle
\begin{abstract}
Semantic type mismatch between a noun and its context is central to coercion phenomena. This paper introduces a graph-based method to examine how lexical and contextual type information is reflected in word embeddings. We select nouns from ten semantic types, annotate corpus instances for type matching (matching vs. coercion vs. other mismatch vs. unrestricted), and construct graphs using BERT and sense-enhanced embeddings. Two metrics—Neighbor Type Probability (NTP) and Neighbor Type Entropy (NTE)—are proposed to analyze neighborhood type distributions. Results show that graphs constructed with sense-enhanced embeddings reflect semantic type information better, and matching and mismatch sentences can be distinguished through the proposed metrics.%
\end{abstract}

\section{Introduction}
Word meaning is composed of various aspects, and semantic type is a fundamental part of it (e.g. \citealp{montague1970universal}). It is related not only to the conceptual category of the noun but also its usage in the grammar. Ideally, semantic types %
can be defined by the distribution of words \citep{asher2011lexical}. For example, the noun `pizza' in \ref{ex:match} belongs to the semantic type \type{food} and typically co-occurs with predicates such as `delicious' and `eat'. %

\ex.\label{ex:introduction}
\a.\label{ex:match}I am eating a delicious pizza.
\b.\label{ex:mismatch}I finished the pizza.

However, nouns are not always used in their prototypical, literal way. %
The context of an instance may require a semantic type different from that of the noun itself. Coercion \citep{pustejovsky1993type} is a typical case of mismatch between the lexical type, i.e. the semantic type of an instance itself and the context type, i.e. the semantic type the context requires or suggests.
In a coercion sentence like \ref{ex:mismatch}, the noun co-occurs with `finish', which typically selects for an \type{activity} rather than a \type{food}. %

In this paper, we investigate how the information about semantic types is reflected in the contextualized word embeddings of noun instances, both in typical cases of normal predication as in \ref{ex:match} and in non-canonical cases of coercion as in \ref{ex:mismatch}. For this purpose, we conduct a graph-based analysis of contextualized word embeddings using cosine similarity between the embeddings to connect instances similar to each other. 

We construct a dataset with nouns from ten semantic types and annotate their corpus occurrences as different types of sentences. Four possible types of sentences are distinguished, with matching sentences and coercion sentences being the main focus of our study. A sentence is seen as a matching sentence if the lexical type matches the contextual type, and coercion is a typical case of mismatch between the lexical type and the contextual type. 

We experiment with different language model embeddings: 
BERT embeddings \citep{devlin-etal-2019-bert} and sense-enhanced embeddings \citep{yavas-etal-2025-relation}. The latter is a fine-tuned variant of BERT in order to incorporate WordNet supersense information \citep{fellbaum_wordnet_1998} into the embeddings. WordNet supersenses correspond well to different semantic types. In addition, we also create graphs using masked versions of these embeddings, where the target word is replaced with a \textsc{[MASK]} token. This allows the focus on the information about contextual types by removing lexical information. We propose two metrics (Neighbor Type Probability ($NTP$) and Neighbor Type Entropy ($NTE$)) to quantify the distribution and diversity of semantic types among each instance’s neighbors in the graph.%

By comparing the neighbor types across the instances within the same lexical type or the same sentence type, we discover that graphs constructed with sense-enhanced embedding reflect the semantic types better than the ones with BERT embeddings; the lexical types are reliably reflected by the graphs constructed with sense-enhanced embeddings, while contextual types are also partially reflected by graphs constructed with masked embeddings.
Instances in different types of sentences display a different pattern in terms of the types of their neighbors.
In matching sentences like \ref{ex:match}, the instances usually share the same lexical type as their neighbors, while in coercion sentences like \ref{ex:mismatch} the instances exhibit a higher diversity of types among the neighbors. These findings indicate the effectiveness of our graph-based method. %

\section{Related work}

\subsection{Coercion}
Coercion has been a key challenge to the theory of semantic types, as the lexical type of a noun in a coercion construction conflicts with the expected types from the context. A range of formal frameworks have been proposed to address this challenge, including Generative Lexicon (GL) \citep{pustejovsky1998generative}, Type Compositional Logic (TCL) \citep{asher2015types}, Modern Type Theory (MTT) \citep{luo2012formal}, frame semantics \citep{long-etal-2022-frame}, among others.

These approaches can be broadly divided into two groups according to how they model the mechanisms underlying coercion. The first group, which includes GL and MTT, assumes a shift in the semantic type of the coerced instance itself. In GL, for example, a noun instance in coercion undergoes a type shift, as in \ref{ex:coercion-in-data}, the type of the noun `bottle' shifted from \type{artifact} to \type{activity}.

\ex.\label{ex:coercion-in-data}I finished the bottle off in two gulps.

The second group, including TCL and frame semantics, assumes a different semantic type not on the coerced noun but on the predication relation, or more generally speaking, the context around the noun. Within frame semantics, for instance, in \ref{ex:coercion-in-data}, the noun \type{bottle} retains its type \type{artifact}, and it is the predicate that accepts an \type{artifact} as its object and creates an eventive reading.
Despite the richness of these theoretical proposals, computational work that directly supports or operationalizes them remains comparatively scarce. One notable exception is \citet{asher-etal-2016-integrating}, who applied distributional models to adjective–noun composition—a domain that includes coercive cases—and provided evidence in favor of TCL.

\subsection{Pre-Trained Language Models and Semantic Types and Coercion}

Few studies focus on whether semantic type knowledge is encoded in the embeddings of pre-trained language models. These studies train classifiers using the frozen embeddings of pre-trained language models and show high performance \citep{zhao-etal-2020-quantifying, yavas-etal-2023-identifying}.

Several studies focus on coercion interpretation with pre-trained language models based on their word predictions and more specifically focusing on coercion to \textit{event}. 
\citet{rambelli-etal-2020-comparing} investigate covert event retrieval in cases of coercion in English, evaluating several pre-trained language models 
against human judgments. Their results show that the models fail to substantially outperform simpler distributional models.

\citet{gietz-beekhuizen-2022-remodelling} show that coercion interpretation is not necessarily resolved to a single covert event. This is evidenced by low human annotator consensus on the underlying event for naturally-occurring coercion sentences in English. They test different computational models including BERT, co-occurrence counts, prototype vector, and example-based learning models. BERT's performance is tested using masked word prediction and it outperforms other computational models in both high and low consensus cases. \citet{ye2022interpreting} evaluate coercion interpretation in English using masked word prediction, measuring whether the top-1 and top-3 predictions of BERT match the underlying covert event and they report poor performance.

\citet{radaelli-etal-2025-compositionality} investigate covert event interpretation in coercion sentences in Norwegian using word prediction across 17 language models. They evaluate whether in models' predictions plausible events are ranked above less plausible ones. Results show that models fail to systematically and consistently rank plausible events higher, and only few outperform a simple corpus frequency baseline.
\citet{radaelli-etal-2025-context}, following \citet{radaelli-etal-2025-compositionality}, investigate how additional context affects coercion interpretation. While the models generally benefit from additional context, the performance varies depending on the model. Models that struggle in context-neutral sentences show greater improvements.

These studies investigate coercion focusing on word predictions of the pre-trained language models. To our knowledge, our study is the first to focus on this phenomenon on the representation level. Furthermore, these studies focus on one type of coercion. We examine a broader range of coercion types and investigate how semantic type relations are more generally captured by the contextualized word embeddings of BERT.

\section{Data}\label{sec:data}
\begin{figure}
\begin{tikzpicture}[scale=0.59]
\Tree[.Entity [.Object [.Animate animal human ] [.Physical\_object artifact food ] [.Abstract\_object info location ] ] [.Event [.Motional\_Event activity process ] [.Static\_Event state mood ] ] ]
\end{tikzpicture}
\caption{\label{fig:type_hierarchy}A presumed type hierarchy of the selected semantic types.}
\end{figure}
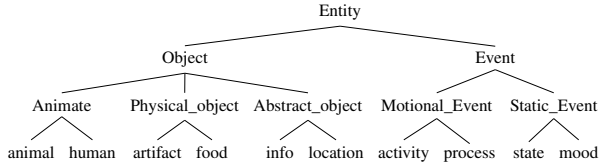

We select a set $\mathcal{T}$ of ten semantic types for analysis: $\mathcal{T}=$~\{\type{animal}, \type{artifact}, \type{activity}, \type{food}, \type{human}, \type{info}, \type{location}, \type{mood}, \type{process}, \type{state}\}. Some of them are ontologically related and can be grouped into higher-level concepts. For example, \type{human} and \type{animal} together form the superordinate category of \type{animate} entities, while \type{state} and \type{mood} can be subsumed under \type{static event}. A conceptual hierarchy of these ten semantic types is shown in Fig.~\ref{fig:type_hierarchy}. These types are selected because their prototypical instances are relatively easy to distinguish. Moreover, each type is presumably close (in the ontological sense) to some types but not others, which motivates a basic hierarchical structure. This hierarchy is broadly compatible to some existing taxonomies of the relevant types (e.g. WordNet \citep{fellbaum_wordnet_1998}) and could be tested with our method.

For each type, we select five to ten relatively frequent nouns. Some types have fewer nouns due to the limited number of high-frequency examples.\footnote{For example, many nouns associated to the type \type{info} tend to be also related to \type{artifact}; however, this study focuses exclusively on nouns with only one single type.} For each selected noun, we randomly extract 20 sentences containing the noun from BookCorpus \citep{zhu2015aligning}. A subset of sentences is manually removed for two reasons. First, some sentences exhibit a similar syntactic and semantic distribution to other existing sentences. Second, certain selected nouns are polysemous and the instance in the extracted sentence corresponds to another sense entry. After the filtering process, each noun is represented by 10 to 16 sentences. All the selected nouns can be found in the Table~7 
and \ref{tab:full-comparison1} in the appendix, and the full dataset will be published upon acceptance.

Each sentence is annotated with information about the semantic types related to the selected noun. Two notions of semantic types are distinguished: lexical type and contextual type (abbreviated as $lt$ and $ct$). Lexical type refers to the semantic type of the noun itself, and contextual type refers to the semantic type that the surrounding context requires or suggests. In practice, the contextual type of an instance can be inferred by masking the noun in the sentence. Consider example \ref{ex:normal}, where the lexical type of the noun `conference' is \type{activity}. When the noun is masked, as in \ref{ex:mask-normal}, the noun that can felicitously fill the masked position can be `meeting', `match', `argument', all of which are associated to \type{activity}. Therefore, the contextual type of `conference' in \ref{ex:normal} is also \type{activity}. Such sentences, where $lt=ct$, are annotated as \textsc{matching}. %
In our dateset, 88\% of the sentences (1410 sentences) are labeled as \textsc{matching}, consistent with the general assumption that in a sentence, the contextual type matches the lexical type.

\ex.\label{ex:normals}
\a.\label{ex:normal}The \textbf{conference} with James Stickley lasted for more than an hour. {\tiny ($lt$: \type{activity})}
\b.\label{ex:mask-normal}The [MASK] with James Stickley lasted for more than an hour.

In addition to the matching cases discussed above, we also encounter instances where $lt\neq ct$, i.e. the semantic type of the noun does not align with the type required or implied from the context. Coercion is a typical case of such a mismatch. In a coercion sentence, the noun combines with a predicate that typically selects for another semantic type. \ref{ex:coercion} is an example of \textsc{coercion}, where the verb `roar' conventionally takes an instance of \type{human} instead of a \type{location} as its subject. Consequently, in \ref{ex:coercion}, the contextual type ($ct$) of `stadium' is \type{human}, while its lexical type ($lt$) is \type{location}, resulting in a clear mismatch. These cases are annotated as \textsc{coercion}. In our dataset, 81 sentences receive this label.

Mismatch between $lt$ and $ct$ can also arise from other mechanisms, such as metaphor (e.g.\ref{ex:metaphor}) and metonymy, though such instances are comparatively rare in our sample. We annotate them under the label \textsc{other\_mismatch}. Sixteen sentences in the dataset are labeled as \textsc{other\_mismatch}. When these two mismatch types are discussed together without distinction, we refer to them collectively as \textsc{mismatch}.

\ex.\label{ex:annotations}
\a.\label{ex:coercion}One moment the \textbf{stadium} was roaring and the next, everything went completely silent. {\footnotesize ($lt$: \type{location}, $ct$: \type{human})}
\b.\label{ex:metaphor}Just because I'm not a social \textbf{butterfly} doesn't mean I'm not smart or capable. {\footnotesize ($lt$: \type{animal}, $ct$: \type{human})}

For all sentences labeled as \textsc{coercion} or \textsc{other\_mismatch}, we additionally annotate the contextual type.\footnote{Lexical types are already known from the noun selection process and therefore do not require re-annotation.}

Besides the above two cases, there is a third situation that the context is highly general and imposes little selectional restriction on the semantic type of the noun. In such cases we say $ct=\emptyset$. Such contexts are theoretically compatible with almost any types of noun. In example \ref{ex:unspecified}, `linguist' can be replaced by any other types of noun such as `stadium', `butterfly', or `conference' without giving rise to semantic anomaly. These cases are annotated as \textsc{unrestricted}. Our dataset contains 82 such sentences.

\ex.\label{ex:unspecified}Perhaps you just need a \textbf{linguist}.\\{\footnotesize ($lt$: \type{human}, $ct$: $\emptyset$)}

An overview of the annotation labels and their relation with the contextual type is summarized as Table~\ref{tab:label}.

\begin{table}[]
\footnotesize
\begin{tabular}{l|llll}
& matching  & coercion    & other\_mis.      & unrestricted    \\
\hline
Types & $lt=ct$ & $lt\neq ct$ & $lt\neq ct$ & $ct=\emptyset$ \\
\# & 1410 & 81 & 16 & 82
\end{tabular}
\caption{\label{tab:label}The meanings of the annotation labels and the number of sentences with the labels}
\end{table}

\section{Method}

Based on the annotated dataset, we construct graphs in which each node corresponds to an annotated instance of a noun. Edges are established according to the similarity between the embeddings of the instances. For each node, we examine its neighboring nodes with respect to their semantic types and propose quantitative measures to characterize the distribution of neighbor types. By comparing these measures across instances belonging to different sentence categories (i.e. \textsc{matching}, \textsc{coercion}, \textsc{other\_mismatch}, \textsc{unrestricted}), we assess the extent to which the relation between the lexical type and the contextual type of an instance is reflected in the graphs.

\subsection{\label{sec:models}Models}

We experiment with the contextualized word embeddings of two language models: BERT \citep{devlin-etal-2019-bert} (base, uncased) and sense-enhanced BERT \citep{yavas-etal-2025-relation}.
Both models are accessed via Hugging Face\footnote{Models: \textit{google-bert/bert-base-uncased} and \textit{yavasde/sense-enhanced-bert}. Via: \url{https://huggingface.co/}} and the Transformers library \citep{wolf-etal-2020-transformers}. Sense-enhanced BERT is a variant of BERT created by fine-tuning in order to incorporate external semantic knowledge into the model embeddings. It is created by fine-tuning BERT on the \mbox{SemCor} corpus \citep{miller-etal-1993-semantic} with WordNet supersense labels. This model is relevant to our study since supersenses are broad semantic categories that correspond well to semantic types, such as \type{animal}, \type{location}, and \type{state}.

We extract the embeddings of the target words in the sentences from the last 4 layers of the models and average them to obtain one embedding per target word instance. Averaging the last four layers has been previously used for semantics-related tasks \citep{liu-etal-2021-mirrorwic, yavas-etal-2025-relation}.
If the target word is tokenized into subwords by the model tokenizer, we average the embedding of each subtoken before averaging across layers. 

To obtain embeddings without lexical type information, we mask the target word in each sentence using the \textsc{[MASK]} token, as shown in \ref{ex:mask-normal}. We extract the embedding of the \textsc{[MASK]} token and use these masked word embeddings for constructing graphs.

\subsection{Graph construction}

We conduct a graph-based analysis on the embeddings. Concretely, we construct a directed graph $G=(V,E)$ over the instances in the dataset, where $V = \{v_1, \dots, v_n\}$ is the set of nodes representing the $n$ instances. Each instance is represented by a contextual word embedding $\mathbf{emb}_i$.

To ensure that edges reflect conceptual similarity between different word types rather than lexical overlap, we only form edges between instances of different target words. A directed edge $(v_i, v_j) \in E$ is formed if $v_j$ is among the $k$ closest neighbors of $v_i$, determined by the cosine similarity between their embeddings, as in:

$sim_{ij}=\cos(\mathbf{emb}(v_i), \mathbf{emb}(v_j)),$

$E=\{(v_i,v_j)|sim_{ij}$ is among the $k$ highest similarities between $v_i$ and other nodes\}

In this paper we set $k=10$.\footnote{We set $k=10$ without systematic evaluation of other values. Since our analysis focuses on relative comparisons across models and sentence types, we expect our conclusions to hold for other values of \textit{k}.}

As introduced in \ref{sec:models}, we use four different types of embeddings: BERT, sense-enhanced BERT, masked BERT, masked sense-enhanced BERT. The graphs using these four embeddings are referred to as $G_b$, $G_s$, $G_{mb}$ and $G_{ms}$.

\subsection{Neighbor type metrics}

We examine the out-neighborhood of each instance, $\mathcal{N}^+(v_i)$, in terms of their types in order to evaluate how well the neighbors reflect the lexical type and the contextual type of the instance. As introduced in \ref{sec:data}, in our dataset, for each instance $v_i$, its lexical type $lt_i \in \mathcal{T}$ and contextual type $ct_i \in \mathcal{T}\cup \emptyset$\footnote{Theoretically, the contextual type could be a type outside our selected set of type $\mathcal{T}$, but this kind of sentence does not exist in our dataset.} can be inferred from the annotation. For example, the instance `stadium' has a lexical type ($lt$) \type{location} and a contextual type  ($ct$) \type{human} in \ref{ex:coercion}.

We calculate the distributions of semantic types in the neighborhood of each instance using the Neighbor Type Probability ($NTP$). For each type $t \in \mathcal{T}$, $NTP(t, v_i)$ measures the proportion of neighbors of $v_i$ that have type $t$, where $t_j$ denotes the type of neighbor $v_j$:

$$NTP(t, v_i) = \frac{|\{v_j \in \mathcal{N}^+(v_i) \mid t_j = t\}|}{k}$$

Using $NTP$, we can determine how much the neighbors of an instance reflect its lexical type $lt_i$ or its contextual type $ct_i$ by calculating $NTP(lt_i,v_i)$ and $NTP(ct_i,v_i)$. 
We refer to the specific case where $NTP$ is computed for the lexical type $lt$ as the Lexical Neighbor Type Matching Ratio ($NTMR_L = NTP(lt, v)$), and for the contextual type $ct$ as the Contextual Neighbor Type Matching Ratio ($NTMR_C = NTP(ct, v)$).

We propose another metric to measure the diversity of the neighbors in terms of their semantic types: Neighbor Type Entropy ($NTE$). $NTE$ calculates the entropy of $NTP$ distribution.
We define the Neighbor Type Entropy $NTE$ as:

$$NTE(v_i)=-\Sigma_{t\in \mathcal{T}}NTP(t,v_i)\times \log(NTP(t,v_i))$$

For an instance $v_i$, if most of its neighbors have the same type, the $NTE$ value is relatively low; if the distribution of the types of its neighbors are diverse, the $NTE$ value is higher.

In general, if the graph is organized based on semantic type relations, the neighbors of an instance should be of the same (lexical or contextual) type or taxonomically related semantic types. Moreover, if the graph highlights the lexical information of the instances, we expect most neighbors of an instance $v_i$ to share the lexical type $lt_i$, i.e. $NTP(lt_i,v_i)\approx1$. In contrast, if the graph reflects contextual information more than lexical information, we expect $NTP(ct_i,v_i)\approx1$.

Furthermore, we expect the values of both $NTP$ and $NTE$ to vary across different sentence types (\textsc{matching}, \textsc{mismatch} and \textsc{unrestricted}). These values are obtained by averaging $NTP$ and $NTE$ values of instances belonging to each sentence type. %
As to the lexical types in \textsc{matching} sentences, since we avoided polysemous uses of the selected nouns in our dataset, the lexical types of the nouns remain stable. Thus, they are expected to be reliably reflected by the lexical types of the neighbors in the graph. More specifically, the $NTMR_L$ of an instance is expected to be high, and the $NTE$ low.

As to \textsc{mismatch} sentences, %
the lexical type of the instance is different from the contextual type. Given that the majority of the instances in our dataset are \textsc{matching} sentences, we expect the types of the neighbors of the instance to be balanced between the two possible options. In this case $NTMR_C$ is likely to be lower and $NTE$ slightly higher. For the instances \textsc{unrestricted} sentences, the contextual types are unclear, so we expect the contextual types of their neighbors to be more diverse. In this case, the $NTMR_C$ of the contextual type should be low and the $NTE$ should be the highest among all sentence categories.

\section{Result analysis}
\subsection{\textsc{matching} sentences}
The Neighbor Type Probability ($NTP$) is calculated for each instance. Table~\ref{tab:accuracy-coer} reports the average $NTMR$ of the instances across different graphs and sentences types. Since $G_{mb}$ and $G_{ms}$ are constructed to capture the information about concextual types instead of lexical types, the $NTMR_L$ on these graphs are significantly lower than on $G_b$ and $G_s$. Therefore, for \textsc{matching} sentences, our analysis focuses on comparing $G_b$ and $G_s$.

\begin{table}[]
\footnotesize
\centering
\begin{tabular}{llccc}
\hline
\textbf{Graph} & \textbf{Sent. type} & \textbf{$NTMR_L$} & \textbf{$NTMR_C$} & \textbf{$other$} \\
\hline
$G_b$ & Matching        & 0.63 & — & 0.37 \\
     & Coercion       & 0.50 & 0.17 & 0.33 \\
     &  Other\_m.       & 0.54 & 0.23 & 0.23 \\
     & Unrestr. & 0.49 & —     & 0.51 \\
\hline
$G_{mb}$ & Matching        & 0.39 & — & 0.61 \\
             & Coercion       & 0.18 & 0.36 & 0.46 \\
   &  Other\_m.       & 0.09 & 0.47 & 0.44 \\
            & Unrestr. & 0.14 & —     & 0.86 \\
\hline
$G_s$ & Matching        & 0.81 & — & 0.19 \\
            & Coercion       & 0.65 & 0.18 & 0.17 \\
            &  Other\_m.       & 0.50 & 0.41 & 0.08 \\
            & Unrestr. & 0.65 & —     & 0.35 \\
\hline
$G_{ms}$ & Matching        & 0.44 & — & 0.56 \\
               & Coercion       & 0.20 & 0.40 & 0.40 \\
            &  Other\_m.       & 0.11 & 0.55 & 0.34 \\
                   & Unrestr. & 0.18 & —     & 0.82 \\
\hline
\end{tabular}
\caption{\label{tab:accuracy-coer}Average Lexical Neighbor Type Matching Ratio ($NTMR_L$) and Contextual Neighbor Type Matching Ratio ($NTMR_C$) and the proportion of neighbor types other than $lt$ and $ct$ ($other$) of different sentence types.}
\end{table}

\begin{figure*}[t]
\centering
\begin{tikzpicture}[scale=0.8]
\begin{axis}[
    height=7cm,
    view={0}{90},
    xtick={0,1,2,3,4,5,6,7,8,9},
    ytick={0,1,2,3,4,5,6,7,8,9},
    xticklabels={animal,arti.,activity,food,human,info,loc.,mood,proc.,state},
    yticklabels={animal,arti.,activity,food,human,info,loc.,mood,proc.,state},
    ticklabel style={font=\scriptsize},
        title={$G_b$},
]
\addplot3[
    matrix plot*,
    mesh/cols=10,
    point meta=explicit,
    nodes near coords,
    nodes near coords style={font=\tiny, text=white},
] table [meta=C] {
x y C
0 0 83
1 0 1
2 0 0
3 0 5
4 0 10
5 0 0
6 0 0
7 0 0
8 0 0
9 0 0
0 1 8
1 1 47
2 1 2
3 1 10
4 1 6
5 1 11
6 1 5
7 1 0
8 1 9
9 1 1
0 2 0
1 2 0
2 2 92
3 2 2
4 2 0
5 2 0
6 2 1
7 2 1
8 2 3
9 2 1
0 3 4
1 3 3
2 3 6
3 3 86
4 3 0
5 3 0
6 3 0
7 3 0
8 3 0
9 3 0
0 4 17
1 4 9
2 4 0
3 4 1
4 4 48
5 4 3
6 4 12
7 4 1
8 4 5
9 4 4
0 5 1
1 5 7
2 5 1
3 5 0
4 5 3
5 5 68
6 5 1
7 5 1
8 5 5
9 5 13
0 6 2
1 6 6
2 6 8
3 6 0
4 6 7
5 6 1
6 6 67
7 6 0
8 6 8
9 6 0
0 7 0
1 7 0
2 7 16
3 7 0
4 7 1
5 7 9
6 7 4
7 7 49
8 7 7
9 7 14
0 8 1
1 8 8
2 8 19
3 8 1
4 8 6
5 8 11
6 8 9
7 8 5
8 8 33
9 8 9
0 9 0
1 9 0
2 9 8
3 9 0
4 9 4
5 9 31
6 9 1
7 9 15
8 9 16
9 9 25
};
\end{axis}
\end{tikzpicture}
\begin{tikzpicture}[scale=0.8]
\begin{axis}[
    colorbar,
    height=7cm,
    view={0}{90},
    xtick={0,1,2,3,4,5,6,7,8,9},
    ytick={0,1,2,3,4,5,6,7,8,9},
    xticklabels={animal,arti.,activity,food,human,info,loc.,mood,proc.,state},
    yticklabels={animal,arti.,activity,food,human,info,loc.,mood,proc.,state},
    ticklabel style={font=\scriptsize},
    title={$G_s$},
]
\addplot3[
    matrix plot*,
    mesh/cols=10,
    point meta=explicit,
    nodes near coords,
    nodes near coords style={font=\tiny, text=white},
] table [meta=C] {
x y C
0 0 86
1 0 4
2 0 0
3 0 2
4 0 7
5 0 0
6 0 1
7 0 0
8 0 0
9 0 0
0 1 3
1 1 79
2 1 0
3 1 10
4 1 0
5 1 0
6 1 4
7 1 0
8 1 2
9 1 0
0 2 0
1 2 1
2 2 88
3 2 2
4 2 0
5 2 0
6 2 2
7 2 1
8 2 6
9 2 0
0 3 1
1 3 6
2 3 1
3 3 91
4 3 1
5 3 1
6 3 0
7 3 0
8 3 0
9 3 0
0 4 6
1 4 5
2 4 1
3 4 1
4 4 86
5 4 0
6 4 2
7 4 0
8 4 0
9 4 0
0 5 0
1 5 0
2 5 0
3 5 0
4 5 0
5 5 85
6 5 1
7 5 1
8 5 0
9 5 13
0 6 0
1 6 11
2 6 1
3 6 0
4 6 1
5 6 0
6 6 84
7 6 0
8 6 2
9 6 0
0 7 0
1 7 0
2 7 0
3 7 0
4 7 0
5 7 13
6 7 1
7 7 61
8 7 1
9 7 24
0 8 0
1 8 1
2 8 17
3 8 0
4 8 1
5 8 0
6 8 3
7 8 4
8 8 70
9 8 5
0 9 0
1 9 0
2 9 3
3 9 0
4 9 1
5 9 22
6 9 3
7 9 15
8 9 12
9 9 44
};
\end{axis}
\end{tikzpicture}
\caption{\label{fig:literal-neighbors} Neighbor Type Ratio ($NTP$) for \textsc{matching} sentences per semantic type, from the graphs based on BERT ($G_b$) (left) and sense-enhanced BERT embeddings ($G_s$) (right). 
The diagonal grids correspond to the average Lexical Neighbor Type Matching Ratio ($NTMR_L$) for each type.}
\end{figure*}
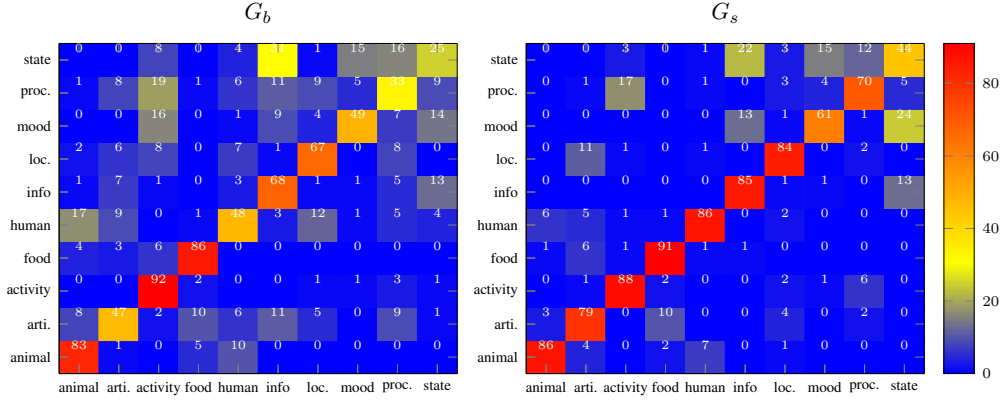

Fig.~\ref{fig:literal-neighbors} presents the average $NTP$ of instances in each lexical type in \textsc{matching} sentences.\footnote{More detailed results are given in the appendix.} 
Both heatmaps indicate that the lexical types of most neighbors coincide with those of the instances themselves. Specifically, on $G_b$, instances of types \type{animal}, \type{activity} and \type{food} exhibit an average $NTMR_L$ over 80\%. This proportion is relatively lower for instances of \type{human}, \type{artifact} and \type{mood}, where fewer than half of their neighbors share the same lexical types.
In contrast, on $G_s$, the average $NTMR_L$ increases for almost all types, most notably for types \type{artifact} and \type{human}. These observations suggest that $G_s$ reflects the lexical types of the instances more faithfully than $G_b$.

A closer investigation of the $NTP$ for \type{human} instances further illustrates the increase of $NTMR_L$ on $G_s$. According to Table~\ref{tab:model-comparison}, certain words, such as `student' and `passenger', have very low $NTMR_L$ values on $G_b$. In particular, most neighbors of `student' instances are of type \type{location} (predominantly `classroom' and `school'), while most neighbors of `passenger' instances are instances of \type{artifact} (mainly `bus' and `truck'). This indicates that $G_b$ reflects not only semantic type of nouns but also co-occurrence information, whereas on $G_s$, the semantic type information is more prominently highlighted.

\begin{table}[]
\centering
\begin{tabular}{l|c|c}
\textbf{Word} & \textbf{$G_b$} & \textbf{$G_s$} \\
\hline
linguist   & 80.6 & 100.0 \\
programmer & 48.1 & 100.0 \\
student    &  9.3 & 100.0 \\
genius     & 54.0 &  99.3 \\
lady       & 40.6 &  99.38\\
terrorist  & 50.6 &  92.6 \\
teenager   & 88.6 & 100.0 \\
coward     & 63.0 &  81.5 \\
passenger  & 15.0 &  43.7 \\
german     & 86.6 &  91.6 \\
\end{tabular}
\caption{Average Lexical Neighbor Type Matching Ratio ($NTMR_L$) of \type{human} type words on $G_b$ and $G_s$}
\label{tab:model-comparison}
\end{table}

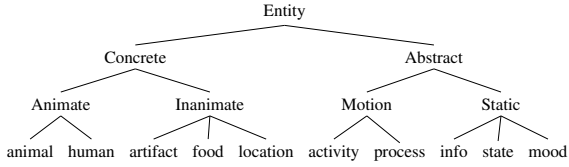
\begin{figure}
\begin{tikzpicture}[scale=0.59]
\Tree[.Entity [.Concrete [.Animate animal human ] [.Inanimate artifact food location ] ] [.Abstract [.Motion activity process ] [.Static info state mood ] ] ]
\end{tikzpicture}
\caption{\label{fig:type_hierarchy_new}A new type hierarchy induced from the $NTP$ values of the semantic types.}
\end{figure}

The $NTP$ computed on $G_s$ also reveals the distances between semantic types. As shown in Fig.~\ref{fig:literal-neighbors}, the $NTP$ between certain pairs of types is relatively higher than between others, indicating their taxonomic proximity. For example, the type \type{artifact} is close to \type{food} and \type{location}; \type{activity} is close to \type{process}; \type{process}, \type{state} and \type{mood} form a cluster; and interestingly, \type{info} is close to \type{mood}. Although this is less common in linguistic studies on type hierarchies, our result suggests a potential prominence in the distinction of abstract and concrete entities. Although this observation requires further investigation, based on the observed $NTP$ patterns, we can build a new semantic type hierarchy, presented in Fig.~\ref{fig:type_hierarchy_new}.

Furthermore, the $NTP$ calculated on $G_s$ can also suggest the types of some non-typical instances. For example, the ontological status of `alien' and `robot' is not straightforward. Using our method, according to Table \ref{tab:alien}, it can be inferred that \type{alien} is a non-typical member of the type \type{human}, and \type{robot} is a non-typical member of \type{artifact}, and they are close to each other and both similar to the type \type{animal}. %

\begin{table}[]
\small
\begin{tabular}{c|cccccc}
      & alien & robot & anim. & arti. & hum. & other \\
      \hline
alien & /     & 0.27   & 0.28    & 0      & 0.39   &  0.06     \\
robot & 0.39   & /     & 0.24    & 0.34      & 0   &  0.03
\end{tabular}
\caption{\label{tab:alien}The neighboring words or types of `alien' and `robot' and their distributions.}
\end{table}

\subsection{\textsc{unrestricted} sentences}
For reasons similar to those observed for \textsc{matching} sentences, the $NTMR_L$ values of instances in \textsc{unrestricted} sentences are higher when computed with unmasked models and with sense‑enhanced models.
Compared with \textsc{matching} sentences, however, the $NTMR_L$ values in \textsc{unrestricted} sentences are lower across all four graphs. The decrease is more significant with masked models, suggesting that the greater diversity of possible contextual types in \textsc{unrestricted} sentences is reflected more clearly on $G_{mb}$ and $G_{ms}$.

The $NTE$ further captures the distinction between \textsc{matching} and \textsc{unrestricted} sentences. Across all four graphs, the diversity of neighbor type distributions is significantly higher for \textsc{unrestricted} sentences than for \textsc{matching} sentences.

\subsection{\textsc{mismatch} sentences}
The results in Table~\ref{tab:accuracy-coer} show that across all graphs, the $NTMR_L$ values of \textsc{mismatch} sentences (both \textsc{coercion} and \textsc{other\_mismatch}) are lower than those of \textsc{matching} sentences. This indicates that the different contextual types exert a noticeable influence on the organization of the graphs.

As discussed in the previous section, the graphs $G_b$ and $G_s$ largely reflect the information about lexical types. On these graphs, most neighbors of the instances in \textsc{coercion} sentences share the same lexical types instead of the contextual types. This finding seems to suggest that instances in \textsc{coercion} sentences retain their lexical types and do not undergo a type shift. A more in-depth research is needed to confirm this observation.

With respect to $NTMR_C$, the graphs constructed with masked models produce higher values than the others. As noted above, in the graphs constructed with unmasked models, especially with sense-enhanced bert, the neighbors of an instance usually have the same lexical type. Consequently, the contextual types of \textsc{mismatch} sentences are underrepresented in these graphs, resulting in $NTMR_C<NTMR_L$.

Table~\ref{tab:CHEN} reveals that no significant difference is found between \textsc{coercion} and \textsc{other\_mismatch} sentences. Therefore, due to the higher number of instances in the dataset, we mainly focus the analysis of the result on \textsc{coercion} sentences.

A comparison of $NTE$  between \textsc{coercion} and \textsc{unrestricted} sentences reveals a pattern distinct from that observed between \textsc{matching} and \textsc{coercion} sentences. On $G_{mb}$ and $G_{ms}$, the $NTE$ of \textsc{coercion} sentences is significantly lower than that of \textsc{unrestricted} sentences, whereas on $G_b$ and $G_s$, no significant difference is found between the two sentence types. This implies that contextual type information is reflected on $G_{mb}$ and $G_{ms}$ to some extent.

For $G_{mb}$ and $G_{ms}$, although $NTMR_C$ is higher than $NTMR_L$, the average value remains lower than 0.5, indicating that the information about contextual types is not highly prominent in these graphs.
This observation is further supported by the analysis of $NTE$. According to Table~\ref{tab:CHEN}, the $NTE$ of instances in \textsc{matching} sentences is significantly lower than the $NTE$ of \textsc{coercion} sentences on $G_b$ and $G_s$, but not on $G_{mb}$ and $G_{ms}$. This suggests that $G_{mb}$ and $G_{ms}$ only partially reflect the information about contextual types. We observe that the graphs sometimes capture other information beyond semantic types, such as morphological clues and co-occurrence patterns. For example, on $G_{mb}$ and $G_{ms}$, the neighbors of the instances of `elephant' are often instances of `omelette' and `explosion', possibly due to the fact that they are all followed by the article `an'.%

Moreover, the information of lexical types is not completely eliminated from $G_{mb}$ and $G_{ms}$. In these two graphs, the $NTP$ of the lexical types are still higher than the $NTP$ of other types. This suggests that the lexical information may persist in \textsc{coercion} sentences. For example, in example \ref{ex:coercion-in-data}, although `finish' strongly indicates a contextual type \type{activity}, the other words in the sentence, such as `gulp', still implies the existence of an \type{artifact} bottle.

\begin{table}[t]
\footnotesize
\centering
\begin{tabular}{llccc l}
\hline
\textbf{G}  & \textbf{mat.} & \textbf{coer.} & \textbf{other.} &  \textbf{unres.} & \textbf{Comparison} \\
\hline
$G_b$  & 0.96 & 1.14 & 0.99 & 1.16 & mat. $<^{**}$ coer. \\
     &          &       &       &       & mat. $<^{**}$ unres. \\
     &          &       &       &       & coer. $\not<$ unres. \\
     &          &       &       &       & coer. $\neq$ other. \\
\hline
$G_{mb}$ & 1.71 & 1.67 & 1.64 & 1.96 & mat. $\not<$ coer. \\
     &          &       &       &       & mat. $<^{***}$ unres. \\
     &          &       &       &       & coer. $<^{*}$ unres. \\
     &          &       &       &       & coer. $\neq$ other. \\
\hline
$G_s$ & 0.61 & 0.85 & 0.65 & 0.83 & mat. $<^{***}$ coer. \\
            &          &       &       &       & mat. $<^{***}$ unres. \\
            &          &       &       &       & coer. $\not<$ unres. \\
                 &          &       &       &       & coercion $\neq$ other. \\
\hline
$G_{ms}$ & 1.45   & 1.49 & 1.42 & 1.77 & mat. $\not<$ coer. \\
            &          &       &       &       & mat. $<^{***}$ unres. \\
            &          &       &       &       & coer. $<^{**}$ unres. \\
                 &          &       &       &       & coer. $\neq$ other. \\
\hline
\end{tabular}
\caption{\label{tab:CHEN}Mean Neighbor Type Entropy ($NTE$) of different sentence types and the statistical comparisons between pairs (Mann-Whitney U test) . $^{*}p < .05$, $^{**}p < .01$, $^{***}p < .001$. $\not<$ indicates the hypothesis was not confirmed. The sentence types \textsc{matching}, \textsc{coercion}, \textsc{other\_mismatch} and \textsc{unrestricted} are shortened as \textsc{mat.}, \textsc{coer.}, \textsc{other.} and \textsc{unres.} respectively.}
\end{table}

In summary, the three types of sentences and their relations with $NTE$ values across the four graphs are summarized as Table~\ref{tab:nte-compare}. This can potentially help in detecting coercion instances.

\begin{table}[]
\centering
\small
\begin{tabular}{c|cc}
             & $NTE$ on $G_b$/$G_s$ & $NTE$ on  $G_{mb}$/$G_{ms}$ \\
             \hline
matching     & low                  & low                         \\
coercion     & high                 & low                         \\
unrestr. & high                 & high                       
\end{tabular}
\caption{\label{tab:nte-compare}The relative numerical relations of Neighbor Type Entropy ($NTE$) values among different types of sentences.}
\end{table}

\section{Conclusive remarks}
This paper has presented a graph‑based analysis of the contextualized word embeddings of noun instances, examining how semantic type information is reflected across different sentence types. We selected nouns from ten semantic types, extracted corpus instances for each noun, and annotated them with lexical types and contextual types. Instances in which the lexical type coincides with the contextual type are labeled as \textsc{matching}; instances where the two diverge are labeled as \textsc{mismatch}, further classified as \textsc{coercion} or \textsc{other\_mismatch}; and instances occurring in uninformative contexts are labeled as \textsc{unrestricted}.

Using these annotated instances, we constructed graphs based on the similarity between their embeddings and proposed two graph‑based metrics in order to analyze the neighbors of instances based on their semantic types: Neighbor Type Probability ($NTP$) and Neighbor Type Entropy ($NTE$). Our results demonstrate that graphs constructed with sense-enhanced embedding reflect information about semantic types better than BERT embeddings. Lexical type information is reliably reflected in graphs constructed on sense-enhanced embeddings, and contextual type information is also partially reflected in graphs constructed on masked embeddings. \textsc{matching}, \textsc{mismatch}, and \textsc{unrestricted} sentences can be distinguished from one another through the comparison of $NTP$ and $NTE$ values across the graphs.

Our method also has a number of potential future applications. On the linguistic side, the distinction between coercion and other mechanisms underlying lexical‑contextual type mismatch can be further explored, given sufficient number of instances with metaphor, metonymy or other kinds of mismatch. A preliminary new type hierarchy has been implied from our method, and with more selection of nouns and, preferably, larger unannotated data, semantic types and their hierarchical organization can possibly be induced automatically, and \textsc{mismatch} sentences can possibly be consistently detected.

On the computational side, the effectiveness of our graph-based method suggests that it can be used to study the mechanisms by which the lexical and contextual type information are encoded and how they interact with each other.

\bibliography{custom, anthology-2}

\appendix

\section{Appendix}
\label{sec:appendix}

\begin{table}[ht]
\centering
\footnotesize
\begin{tabular}{|l|l|c|c|}
\hline
\textbf{Semantic Type} & \textbf{Word} & \textbf{$G_b$} & \textbf{$G_s$} \\
\hline
\multirow{12}{*}{info}
 & message     & 63.75 & 100.00 \\
 & information & 88.75 &  90.00 \\
 & poem        & 13.13 & 100.00 \\
 & rumor       & 88.75 & 100.00 \\
 & data        & 81.33 &  90.00 \\
 & news        & 100.00 & 100.00 \\
 & idea        & 93.75 &  96.88 \\
 & concept     & 89.38 & 100.00 \\
 & science     & 23.75 &  61.25 \\
 & knowledge   & 78.13 &  67.50 \\
 & problem     & 10.63 &  20.63 \\
 & secret      & 81.88 &  99.38 \\
\hline
\multirow{6}{*}{mood}
 & happiness  & 31.88 &  70.00 \\
 & empathy    & 76.88 &  66.88 \\
 & resentment & 86.25 &  97.50 \\
 & preference & 46.25 &  75.00 \\
 & attention  & 23.13 &  13.13 \\
 & mood       & 43.75 &  65.63 \\
\hline
\multirow{10}{*}{animal}
 & elephant  & 87.33 &  99.33 \\
 & dinosaur  & 97.50 &  96.88 \\
 & pig       & 82.00 &  79.33 \\
 & cat       & 100.00 &  98.46 \\
 & cow       & 67.33 &  84.00 \\
 & butterfly & 100.00 & 100.00 \\
 & penguin   & 100.00 &  93.85 \\
 & pigeon    & 99.33 &  94.00 \\
 & mammal    & 99.38 & 100.00 \\
 & bug       & 90.91 &  96.36 \\
\hline
\multirow{10}{*}{location}
 & cave      & 99.38 &  98.75 \\
 & classroom & 58.75 &  87.50 \\
 & stadium   & 37.50 &  93.75 \\
 & school    & 88.75 &  89.38 \\
 & river     & 88.75 &  90.00 \\
 & road      & 43.75 &  22.50 \\
 & valley    & 100.00 &  99.38 \\
 & desert    & 95.00 & 100.00 \\
 & sky       & 11.88 &  80.63 \\
 & park      & 60.00 &  95.00 \\
\hline
\multirow{7}{*}{state}
 & death        & 24.38 &  24.38 \\
 & existence    & 26.88 &  79.38 \\
 & chaos        & 21.88 &  46.25 \\
 & difficulty   & 21.25 &  13.75 \\
 & strength     & 63.75 &  91.25 \\
 & beginning    & 16.88 &  53.13 \\
 & intelligence &  7.50 &  20.63 \\
\hline
\end{tabular}
\label{tab:full-comparison}
\caption{The average $NTMR_L$ value of each noun per semantic type on $G_b$ and $G_s$}
\end{table}

\begin{table}[ht]
\centering
\footnotesize

\begin{tabular}{|l|l|c|c|}
\hline
\textbf{Semantic Type} & \textbf{Word} & \textbf{$G_b$} & \textbf{$G_s$} \\
\hline
\multirow{10}{*}{activity}
 & picnic      & 88.00 &  91.33 \\
 & barbecue    & 71.88 &  61.88 \\
 & banquet     & 90.63 &  90.63 \\
 & celebration & 83.75 &  75.00 \\
 & festival    & 98.13 &  97.50 \\
 & carnival    & 96.00 &  94.00 \\
 & conference  & 99.38 &  90.00 \\
 & meeting     & 98.75 &  97.50 \\
 & parade      & 86.67 &  77.33 \\
 & party       & 90.63 &  85.00 \\
\hline
\multirow{10}{*}{process}
 & explosion    & 56.25 &  99.38 \\
 & accident     & 74.38 &  85.63 \\
 & experiment   & 12.50 &  78.13 \\
 & installation & 28.13 &  90.63 \\
 & interruption & 15.00 &  87.50 \\
 & storm        & 20.63 &  68.13 \\
 & sleep        &  0.00 &  12.67 \\
 & dance        & 16.25 &  17.50 \\
 & attack       & 65.00 &  95.63 \\
 & performance  & 23.13 &  53.75 \\
\hline
\multirow{13}{*}{food}
 & pizza     & 91.88 &  88.75 \\
 & kebab     & 100.00 & 100.00 \\
 & rice      & 97.50 & 100.00 \\
 & omelette  & 100.00 & 100.00 \\
 & pork      & 86.88 &  95.63 \\
 & beef      & 100.00 & 100.00 \\
 & bread     & 98.75 & 100.00 \\
 & cake      & 86.88 & 100.00 \\
 & candy     & 73.57 &  92.86 \\
 & cigarette & 16.88 &  10.63 \\
 & apple     & 93.57 & 100.00 \\
 & carrot    & 84.00 &  96.67 \\
 & potato    & 100.00 & 100.00 \\
\hline
\multirow{12}{*}{artifact}
 & bus       & 90.00 & 100.00 \\
 & truck     & 68.75 & 100.00 \\
 & bicycle   & 81.88 & 100.00 \\
 & rocket    & 12.14 &  58.57 \\
 & guitar    & 43.75 &  93.75 \\
 & sculpture &  1.25 &  71.25 \\
 & computer  &  1.88 &  72.50 \\
 & shirt     & 80.63 &  97.50 \\
 & coin      & 25.63 &  48.13 \\
 & cable     & 43.75 &  85.63 \\
 & button    & 77.50 &  99.38 \\
 & bottle    & 69.38 &  68.13 \\
\hline
\multirow{10}{*}{human}
 & linguist   & 80.63 & 100.00 \\
 & programmer & 48.13 & 100.00 \\
 & student    &  9.38 & 100.00 \\
 & genius     & 54.00 &  99.33 \\
 & lady       & 40.63 &  99.38 \\
 & terrorist  & 50.67 &  92.67 \\
 & teenager   & 88.67 & 100.00 \\
 & coward     & 63.08 &  81.54 \\
 & passenger  & 15.00 &  43.75 \\
 & german     & 86.67 &  91.67 \\
\hline
\end{tabular}
\caption{The average $NTMR_L$ value of each noun per semantic type on $G_b$ and $G_s$}
\label{tab:full-comparison1}
\end{table}

\end{document}